\documentclass{article}

\usepackage{times}
\usepackage{graphicx} 
\usepackage{subfigure} 

\usepackage{natbib}

\usepackage{algorithm}
\usepackage{algorithmic}

\usepackage{hyperref}

\usepackage[draft]{icml2017}

\usepackage{amssymb}
\usepackage{amsmath}
\usepackage{amsthm}
\usepackage{verbatim}

\icmltitlerunning{Semi-Supervised Translation with MMD Networks}

\begin{document} 

\twocolumn[
\icmltitle{Semi-Supervised Translation with MMD Networks}



\icmlsetsymbol{equal}{*}

\begin{icmlauthorlist}
\icmlauthor{Mark Hamilton}{yale}
\end{icmlauthorlist}

\icmlaffiliation{yale}{Yale University, New Haven, CT 06520 USA}

\icmlcorrespondingauthor{Mark Hamilton}{mhamilton723@gmail.com}

\icmlkeywords{Semi-supervised Learning, Maximum Mean Discrepency, MMD, Translation, SMT, Regression}

\vskip 0.3in
]




\begin{abstract} 
This work aims to improve semi-supervised learning in a neural network architecture by introducing a hybrid supervised and unsupervised cost function. The unsupervised component is trained using a differentiable estimator of the Maximum Mean Discrepancy (MMD) distance between the network output and the target dataset. We introduce the notion of an $n$-channel network and several methods to improve performance of these nets based on supervised pre-initialization, and multi-scale kernels. This work investigates the effectiveness of these methods on language translation where very few quality translations are known \textit{a priori}. We also present a thorough investigation of the hyper-parameter space of this method on both synthetic data.

\end{abstract} 

\section{Introduction}

Often in data analysis, one has a small set of quality labeled data, and a large pool of unlabeled data. It is the task of semi-supervised learning to make as much use of this unlabeled data as possible. In the low-data regime, the aim is to create models that perform well after seeing only a handful of labeled examples. This is often the case with machine translation and dictionary completion, as it can be difficult to construct a large number of labeled instances or a sufficiently large parallel corpora. However, this domain offers a huge number of monolingual corpora to make high quality language embeddings \cite{dictionary,al2013polyglot}. The methods presented in this paper are designed to take into consideration both labeled and unlabeled information when training a neural network. The supervised component uses the standard alignment-based loss functions and the unsupervised component attempts to match the distribution of the network's output to the target data's distribution by minimizing the Maximum Mean Discrepancy (MMD) ``distance'' between the two distributions. This has the effect of placing a prior on translation methods that preserve the distributional structure of the two datasets. This limits the model space and increases the quality of the mapping, allowing one to use less labeled data. 

Related methods such as Auto-Encoder pre-initialization \cite{erhan2010does}, first learn the structure of the input, then learn a mapping. In this setup, unsupervised knowledge enters through learning good features to describe the dataset. The MMD method of unsupervised training directly learns a mapping between the two spaces that aligns all of the moments of the mapped data and the target data. This method can be used to improve any semi-supervised mapping problem, such as mappings between languages \cite{lt_trans}, image labeling, FMRI analysis  \cite{mitchell2008predicting}, and any other domains where transformations need to be learned between data. This investigation aims to study these methods in the low data regime, with the eventual goal of studying of dying or lost languages, where very few supervised training examples exist.

\section{Background}

\subsection{Maximum Mean Discrepancy}
\label{MMD1}

\renewcommand{\(}{\left(}
\renewcommand{\)}{\right)}
\newcommand{\X}{\mathcal{X}}
\newcommand{\R}{\mathbb{R}}
\newcommand{\F}{\mathcal{F}}
\newcommand{\N}{\mathcal{N}}
\newcommand{\E}[1]{\mathbb{E}_{#1}}

The Maximum Mean Discrepancy (MMD) put forth by \cite{kernel_two_sample} is a measure of distance between two distributions $ p , q $. More formally, letting $x$, $y$ be variables defined on a topological space $ \X$ with Borel measures $p, q$, and $\F$ be a class of functions from  $ \X \to  \R$. The MMD semi-metric is defined as:

\begin{equation}
 MMD_{\F} \( p,q \) = sup_{f \in \F} \Bigl( \E{x \sim p} f\(x\) -  \E{y \sim q} f\(y\)  \Bigr) 
 \end{equation}
 
Where $\mathbb{E}$ is the first raw moment defined as:

\begin{equation}
    \E{x \sim p} f(x) = \int_{ \X } f(x) dp 
\end{equation}

Intuitively, the MMD is a measure of distance which uses a class of functions as a collection of ``trials'' to put the two distributions through. The distributions pass a trial if the function evaluated on both distributions has the same expectation or mean. Two distributions fail a trial if they yield different means, the size of the difference measures how much the distributions fail that trial. Identical distributions should yield the same images when put through each function in $\F$, so the means (first moments) of the images should also be identical. Conversely, if the function class is ``large enough'' this method can distinguish between any two probability distributions that differ, making the MMD a semi-metric on the space of probability distributions. A unit ball in a Reproducing Kernel Hilbert Space (RKHS) is sufficient to discern any two distributions provided the kernel, $ k$, is universal. \cite{cortes} If $\F$ is equal to a unit ball in kernel space, Gretton $\it{et. al.}$ showed that the following is an unbiased estimator of the MMD: \cite{kernel_two_sample}

\begin{multline}
    MMD_u^2(X,Y) =  \frac{1}{m(m-1)} \sum_{i=1}^{m} \sum_{j \neq i}^{m} k(x_i,x_j) +\\
    \frac{1}{n(n-1)} \sum_{i=1}^{n} \sum_{j \neq i}^{n} k(y_i,y_j) -   \frac{2}{mn} \sum_{i=1}^{m} \sum_{j = 1}^{n} k(x_i,y_j)
\end{multline}

If the kernel function is differentiable, this implies that the estimator of the MMD is differentiable, allowing one to use it as a loss function that can be optimized with gradient descent. 

\subsection{MMD Networks}

The differentiability of the MMD estimator allows it to be used as a loss function in a feed-forward network. Li $\it{et. al.}$ showed that by using the MMD distance as a loss function in a neural net, $\N$, one can learn a transformation that maps a distribution of points $ X = (x_{i})_{1}^{n}$ in $\R^{d}$ to another distribution $Y = (y_{i})_{1}^{m}$ in $\R^{n}$ while approximately minimizing the MMD distance between the image of $X$, $\N(X)$, and $Y$. \cite{GMMN}

\begin{equation}
    l_{MMD}(X,Y,\N) = MMD_u^2(\N(X),Y)
\end{equation}

 This loss function allows the net to learn transformations of probability distributions in a completely unsupervised manner. Furthermore, the MMD-net can also be used to create generative models, or mappings from a simple distribution to a target distribution.\cite{GMMN} Where simple usually means easy to sample from, or a maximum entropy distribution. Often, a multivariate uniform or Gaussian source distribution is used in these generative models. This loss function can be optimized via mini-batch stochastic gradient descent, though the samples from X and Y need not be paired in any way. To avoid over-fitting, the minibatches for X and Y should be sampled independently, which this paper refers to as ``unpaired'' minibatching. 

\section{Methods}

\subsection{$n$-Channel Networks}

 This work introduces a generalization of a feed forward net, called an $n$-Channel net. This architecture allows an unsupervised loss term that requires unpaired mini-batching and a paired mini-batching scheme of a standard feed forward network to be mixed.
 
 An $n$-channel net is a collection of $n$ networks with tied weights that operate on $n$ separate datasets $(X_i,Y_i)_1^n$. More formally, an $n$-channel net is a mapping:
 
\begin{equation}
\N_n : \(\R^d\)^n \to \( \R^e \)^n
\end{equation}
 
 defined as:
 
\begin{equation}
\N_n\Bigl(\(X_i\)_1^n\Bigr) \equiv  \Bigl(\N(X_i)\Bigr)_1^n \end{equation}
 
 where where $\N: \R^d \to \R^e$ is a feed forward network. Each channel of the network can have it's own loss function and be fed with a separate data source. Most importantly, these separate data sources can be trained in a paired or unpaired manner. 

\subsection{A Semi-Supervised MMD-Net}

In many applications where one is interested in estimating a transformation between data spaces, one has a small labeled dataset $(X,Y)$, and large, unlabeled datasets $(S,T)$. Throughout the literature, MMD networks have only been applied to the case of unpaired data.\cite{GMMN}  We expand on this work by augmenting the completely unsupervised MMD distance with a semi-supervised alignment term. More formally, if one has a collection of k paired vectors $(x_{i},y_{i})_{1}^{k}$ with $x_{i} \in X$ and $y_{i} \in Y$ that should be aligned through the transformation $\N$, one can use the standard loss function:

\begin{equation}
    l_{alignment}(X,Y,\N) = \sum_{i=1}^{k} \lVert \N(x_{i})-y_{i} \rVert
\end{equation}

Where $\lVert \cdot \rVert$ is any differentiable norm in $\R^d$. This work uses the standard $l_2$ vector norm. This is the standard norm used in regression, where the goal of the network is to minimize the distance between the network output$\N(x_{i})$, and the observed responses $y_{i}$.

Using a hyperparameter, we can blend the cost functions of the supervised alignment loss and the unsupervised MMD loss. The full cost function for the MMD network then becomes:

\begin{multline}
    l(X,Y,S,T,\N) = \alpha_{pair}l_{alignment}(X,Y,\N) +\\
    (1-\alpha_{pair}) l_{MMD}(S,T,\N) 
\end{multline}

\subsection{Supervised Pre-Initialization}

The MMD term of the cost function scales as $\mathcal{O}\(M^2\)$ where $M$ is the size of the mini-batch. This significantly increases training time for large batch sizes slowing convergence in wall-time. To mitigate this effect, we first train the network until convergence with only the supervised term of the cost function. Once converged, we then switch to the semi-supervised cost function.

This also helps the network avoid local minima as it already starts close to the optimal solution. Because the MMD cost function is inherently unpaired, it is susceptible to getting stuck in local minima when there are multiple ways to map the mass of one probability distribution into another distribution. We say that a mapping from the supports, $f: \X \to \mathcal{Y}$, is a MMD-mode from distributions $p$ to $q$  if $f(p) \sim q$. Here $f(p)$ is the distribution formed by sampling from $p$ and then applying $f$. These modes coincide with critical points of the $MMD_u^2$ cost function and are therefore tough to escape with gradient descent methods. As the class of functions represented by the network increases, the more distinct MMD-modes arise. This increases the number of critical points, though these probably tend to be saddle points rather than local minima as the dimensionality of the function space increases. \cite{saddle} 

One can escape these local minima, by increasing $\alpha_{pair}$ to the point where the signal from the supervised term overcomes that the signal from the unsupervised cost function. However, if the network is within the pull of the correct minima, it is often better to rely on the robust unsupervised signal than the noisy supervised signal, which requires a small $\alpha_{pair}$. We found that supervised pre-training helped guide the network parameters to within the basin of attraction for the correct unsupervised minima. From here the unsupervised signal was much more reliable and led to better results on synthetic and language datasets. Furthermore on all datasets, the supervised warm-start greatly reduced fitting time, as convergence of the expensive MMD cost function needed fewer optimization steps. Future work could involve annealing the supervised term to a small number, though this would eliminate the aforementioned computational speedup.  

\begin{figure*}[t!]
    \centering
    \begin{subfigure}
        \centering
        \includegraphics[height=1.6in]{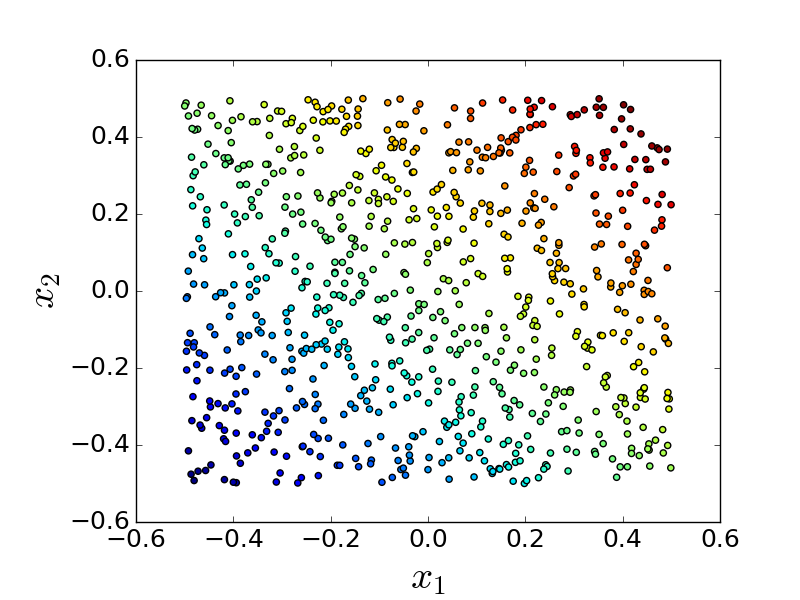}
    \end{subfigure}%
    \begin{subfigure}
        \centering
        \includegraphics[height=1.6in]{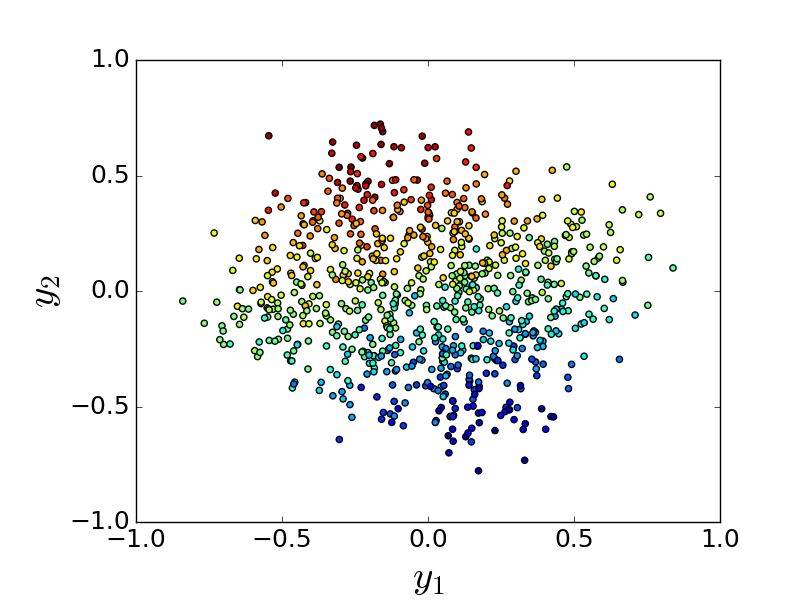}
    \end{subfigure}
    \begin{subfigure}
        \centering
        \includegraphics[height=1.6in]{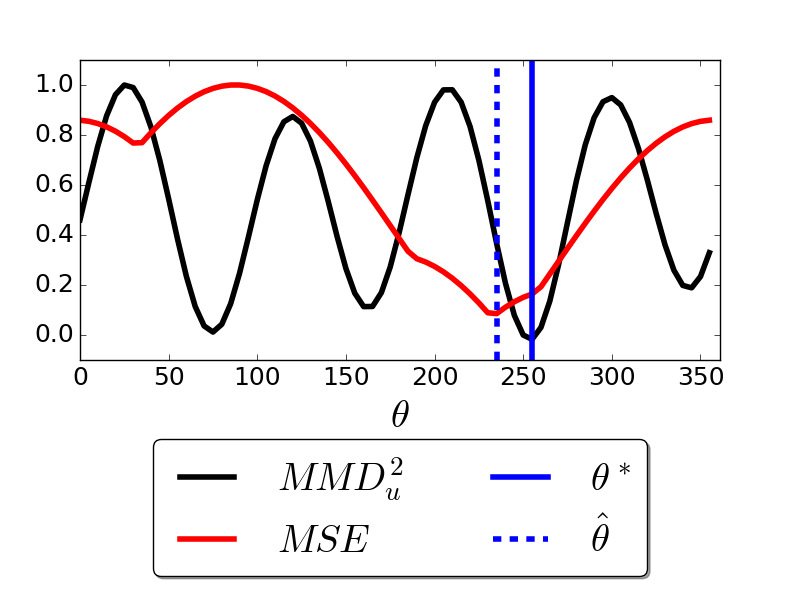}
    \end{subfigure}
    \caption{Left: Initial dataset $X$ sampled uniformly from the unit square. Colors indicate how points are mapped through the transform. Middle: $Y = X_{255^{\circ}} + Gaussian(\mu=0,\sigma=.1) $ Where $X_{\theta}$ denotes a rotation clockwise by $\theta$. Right: Unit scaled $MMD_u^2(X_{\theta},Y)$, and unit scaled $MSE(X_{\theta,1},Y_1)$ as a function of $\theta$. Where $X_{1}$ denotes the first element of $X$.}
     \label{mmd_vs_angle}
\end{figure*}

To demonstrate the effect of pre-initialization, we show the unbiased MMD estimator of a simple synthetic experiment. We generate two datasets of two dimensional points. The first, shown in Figure \ref{mmd_vs_angle} left is sampled from a uniform distribution on the unit square support centered at $(0,0)$. To generate a simple target shown in Figure \ref{mmd_vs_angle} middle, we rotate the source cloud of points by an angle $\theta^*=255^{\circ}$ and add a small Gaussian noise term. Figure \ref{mmd_vs_angle} right shows the that MMD loss as a function of angle of rotation transformation has several modes caused by the symmetries of the square. To simulate a very noisy MSE, we use the MSE of one randomly sampled point and its respective pair. The noisy MSE loss function has two local minima and the global minima $\hat{\theta}$ is within the correct basin of attraction of the unsupervised cost function. This basin of attraction of the unsupervised cost has a minima that is indistinguishable from the correct value of theta and much more accurate than the supervised loss term.

\subsection{Choice of Kernel}

The $MMD_{\F}$ is able to differentiate between any two distributions if the function class, $\F$, is a unit ball in the reproducing kernel Hilbert space (RKHS) of a universal kernel.\cite{cortes} One of the simplest and most commonly used universal kernels is the Gaussian or radial basis function kernel, which excels at representing smooth functions. 

\begin{equation}
k_{\sigma}(x,y) =   exp\(- \frac{\lVert x - y \rVert^2}{2\sigma^2} \)
\end{equation}

The parameter $\sigma$ controls the width of the Gaussian, and needs to be set properly for good performance. If $\sigma$ is too low, each point's local neighborhood will be effectively empty, and the gradients will vanish. If it is too high, every point will be in each point's local neighborhood and the kernel will not have enough resolution to see the details of the distribution. In this scenario, the gradients vanish. We found that $\sigma$ was one of the most important hyper-parameters for the success of the method. In both our synthetic data and natural language examples, we found that the method performed well in a small window of kernel scale settings.

To improve the robustness of this method, this investigation used the following multi-scale Gaussian kernel:

$$ k(x,y) = \sum_{i = 0}^{n} c_i k_{\sigma_i}(x,y) $$

Where  $c_i = 1$, $\sigma_i = s 10^{w(i/n)-w/2}$, $w = 4$, $n=10$. The scalar $s$ is the average scale of the multi-scale and the width, $w$, controls the width of the frequency range covered by the kernel. $n$ controls how many samples are taken from this range. Choosing a larger $n$ improves performance as there are more scales in the kernel, but increases computation time. By including multiple scales in the kernel, the gradients from the larger kernels will move the parameters to a region where the distributions are aligned at a large scale, they will then begin to vanish and the smaller scale gradients will become more relevant. Setting $w=4$ allows the kernel to be sensitive to functions with scales that are within $2$ orders of magnitude of the average scale $s$. We find that choosing this kernel significantly broadens the areas of parameter space where the method succeeds, without hurting the performance. 

\begin{figure*}[t!]
	\centering
	\begin{subfigure}
		\centering
		\includegraphics[height=1.6in]{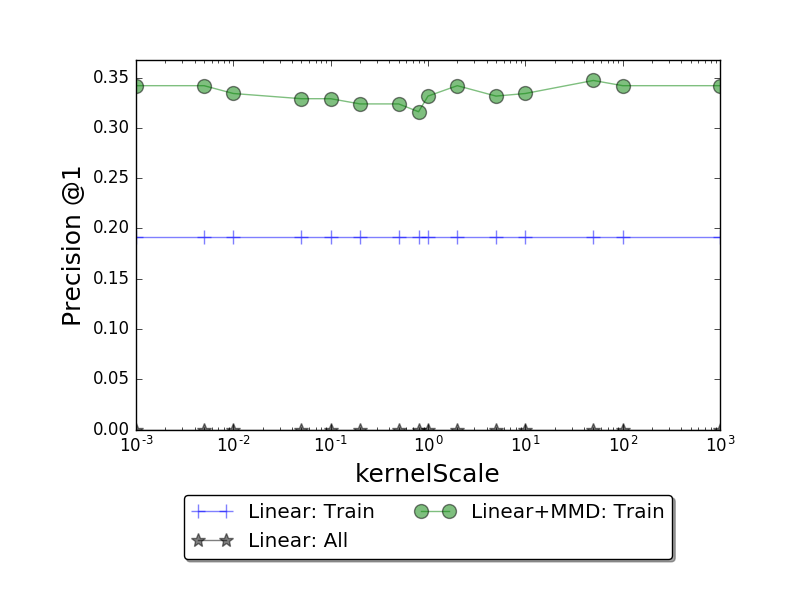}
	\end{subfigure}%
	\begin{subfigure}
		\centering
		\includegraphics[height=1.6in]{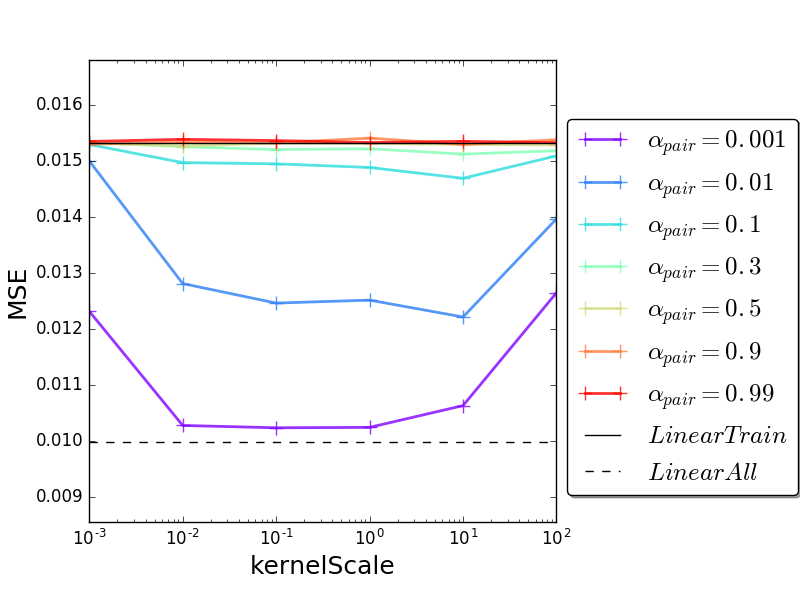}
	\end{subfigure}
	\begin{subfigure}
		\centering
		\includegraphics[height=1.6in]{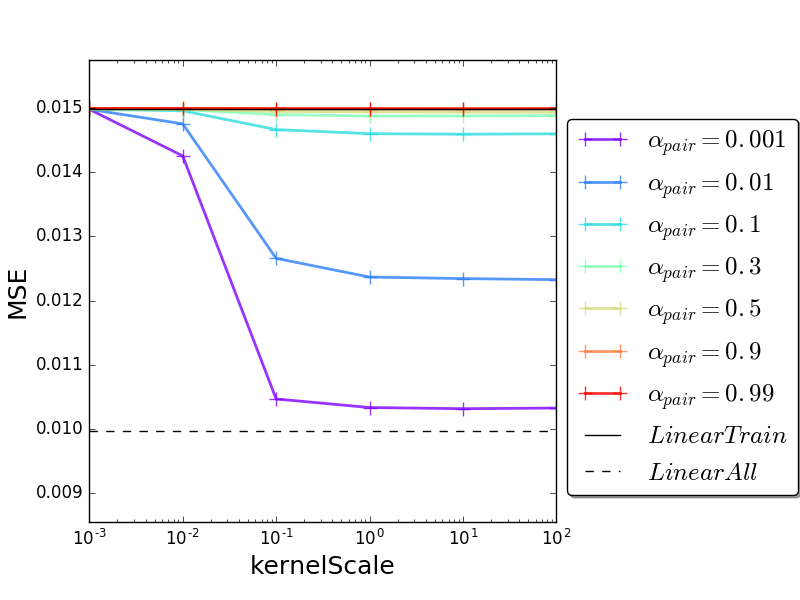}
	\end{subfigure}
	\caption{Left: Performance comparison on word embeddings in the $0-5k$ frequency bin as a function of the average kernel scale $s$. Middle:  Performance comparison on synthetically generated data in $\R^{30}$ as a function of $\alpha_{pair}$. Right:  Performance comparison on synthetically generated data in $\R^{300}$ as a function of $\alpha_{pair}$.}
	\label{kernel_scale}
\end{figure*}

Many have investigated the kernel scale problem and there are several heuristics available for choosing the scale based on optimizing statistical power or median distances to nearest neighbors. \cite{optimal} For clarity, we explicitly investigated and set the kernel scale based on a grid search evaluating on a held out validation set. Figure \ref{kernel_scale} demonstrates that the method was fairly robust to settings of average kernel scale on synthetic data and language data.

\subsection{Globally Corrected (GC) Retrieval}

In this analysis, performance of translation methods are compared on their ability to infer the correct translation on a held out test set. More specifically, we use the precision at $N$, which is the fraction of examples where the correct word was in the top $N$ most likely translations of model. This is a natural choice for translation, as it estimates the probability of translating a word correctly when $N=1$.

To generate the list of $N$ most likely translations for a given word, one can use nearest neighbor (NN) retrieval. In this method, one uses the $N$ closest neighbors in the target space of the mapped word vector as the list of best guesses. We find that it is always better to use cosine distance for nearest neighbor calculations. Finding the first nearest neighbor to a point $\hat{y}$ can be more formally expressed as:

\begin{equation}
NN_1(\hat{y}) = argmin_{y \in T} Rank_T(\hat{y},y)
\end{equation}

Where $\hat{y}$ is our mapped word vector, $T$ is our target space, and $Rank_T(\hat{y},y)$ is a function that returns the rank of $y$ in the sorted list of distances between $\hat{y}$ and the points in $T$. 

If the space of word embeddings is not uniformly distributed, there will be areas where word embeddings bunch together in higher densities. The points towards the center of these bunches act as hub points, and may be the nearest neighbors of many other points. Dinu \textit{et. al.} 2014 have shown that naive NN retrieval results in over-weighting these hub points as they are more frequently the neighbors of points. They called this the ``Hubness Problem'' and introduced a corrected form of the nearest-neighbor retrieval called the globally corrected neighbor retrieval method (GC). In this method, instead of using distance to select translates as in $NN_1$, one uses:

\begin{equation}
GC_1(\hat{y})=argmin_{y \in T} \(Rank_P\(y,\hat{y}\)-cos\(\hat{y},y\)\)
\end{equation}

Where $P$ is a random sampling of points from $T$ and $cos(x,y)$ is the cosine distance between $x$ and $y$. Instead of returning the nearest neighbor of $\hat{y}$, GC returns the point in $T$ that has $\hat{y}$ ranked the highest. The cosine distance term breaks ties. GC retrieval has been shown to outperform the nearest neighbor retrieval in all frequency bins when the transformation is a linear mapping.\cite{lt_trans}  Figure \ref{emb_comp} shows that it also improves the performance of the semi-supervised translation task.

\subsection{Neural Network Implementation}

This work implemented the network in Theano, \cite{theano} an automatic differentiation software written in python. The net was trained with RMSProp \cite{RMSprop} on both the unpaired and paired batches with a batch size of 200 for each set. The unregularized pre-initialization was trained for $4000$ epochs and the regularized network was trained for $250$ epochs, which gave ample time for convergence. Hyperparameter optimization was perfomed through parallel grid searches a TORQUE Cluster, where each job ran for $\sim20$ hours. A validation set consisting of a random sample of $10\%$ of the training set was used to choose the parameters for the final reported results.

\section{Data}

\subsection{Synthetic Data}

Several synthetic datasets were used to demonstrate the method's ability to accurately learn linear transformations using a very small paired dataset. Furthermore, we used this synthetic data to investigate the effects of the network's hyper-parameters.

Two datasets were created, one with the dimension of the source and target equal to $30$ and the other $300$, the same dimensionality as the embeddings. The datasets contained $100,000$ points and various sized paired subsets were used to calculate the supervised alignment loss in the experiments.

Source data was generated as a multivariate Gaussian with zero mean and unit variance. A ground truth mapping was generated by sampling the entries of a $d \times d$ matrix of independent Gaussians with zero mean and unit variance. The target data was generated by applying the ground truth transformation to the source data and adding Gaussian noise with zero mean and a variance of $0.1$.

\subsection{Embedding Data}
This analysis used 300 dimensional English (EN) and Italian (IT) monolingual word embeddings from \cite{lt_trans}. These embeddings were trained with word2vec's CBOW model on $2.8$ billion tokens as input (ukWaC + Wikipedia + BNC) for English and the 1.6 billion
itWaC tokens for Italian.\cite{lt_trans} The embeddings contained the top 200,000 words in each language. Supervised training and testing sets were constructed from  a dictionary built from Europarl, available at \url{http://opus.lingfil.uu.se/}. \cite{dictionary}

Two training sets consisted of the $750$ and $5,000$ most frequent words from the source language (English) which had translations in the gold dictionary. Five disjoint test sets were created consisting of roughly 400 translation pairs randomly sampled from the frequency ranked words in the intervals 0-5k, 5k-20k, 20k-50k, 100k-200k. 

\section{Results}

\subsection{Synthetic Data}

Adding the MMD term to the loss function dramatically improved the ability to learn the transformation on all synthetic datasets. The synthetic data also provided a clean environment to see the effect of varying hyper-parameters. The experiment used a ``linear network'' which is equivalent to learning a linear transformation between the spaces. In general, if the hyper-parameters are set correctly, the MMD assisted learner can approach the true transformation with significantly less  paired data.

Our first investigation aimed to understand the effect and robustness of the kernel scale parameter. As one can see from Figure \ref{kernel_scale}, the performance of the method is robust to a setting of the average kernel scale within $+/-2$ orders of magnitude of the optimal scale. This empirically confirms the intuition behind the width parameter of the multi-scale kernel. As the width parameter decreases, this valley of good performance becomes narrower by the expected amount. A similar pattern arose in the $300$ dimensional dataset.

In order to simulate the environment of the embedding experiment that required a validation set of $\sim 10\%$ of the data, we also removed $\sim 10\% $ of our data. The plots in Figure \ref{kernel_scale} demonstrate that even with the data removed for a validation set, the method still significantly beats linear regression trained on the training and validation set, justifying the use of data for parameter tuning. The models in $d=30$ and $d=300$ both reach error rates comparable to the ground truth regressor learned on all $100,000$ data points.

\begin{figure}
	\includegraphics[height=2.6in]{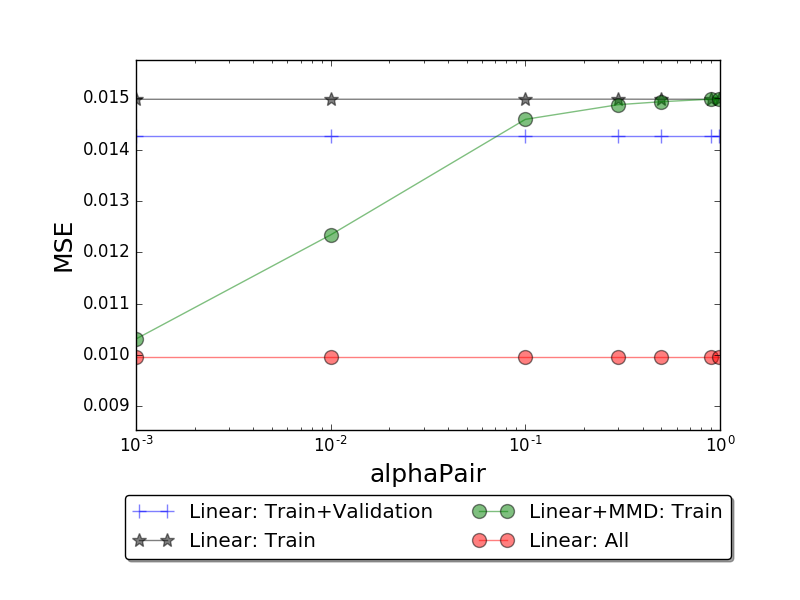}
	\caption{Performance of methods on synthetically generated data in $\R^{300}$ as a function of $\alpha_{pair}$, $s=10$.}
	\label{alpha_pair}
\end{figure}

Figure \ref{alpha_pair} investigates various settings of $\alpha_{pair}$ and shows that decreasing $\alpha_{pair}$ drives the performance down to the ground truth level. This trend appears in both the low and high dimensional data and suggests that the supervised pre-initialization yields a configuration that is within the basin of attraction of the true parameters in vector field $\nabla l_{MMD}$. Thus, only the unsupervised term is needed as the supervised initialization has already eliminated the ambiguity of the MMD loss function modes.

\subsection{Embedding Data}

\begin{figure*}[t!]
    \centering
    \begin{subfigure}
        \centering
        \includegraphics[height=2.5in]{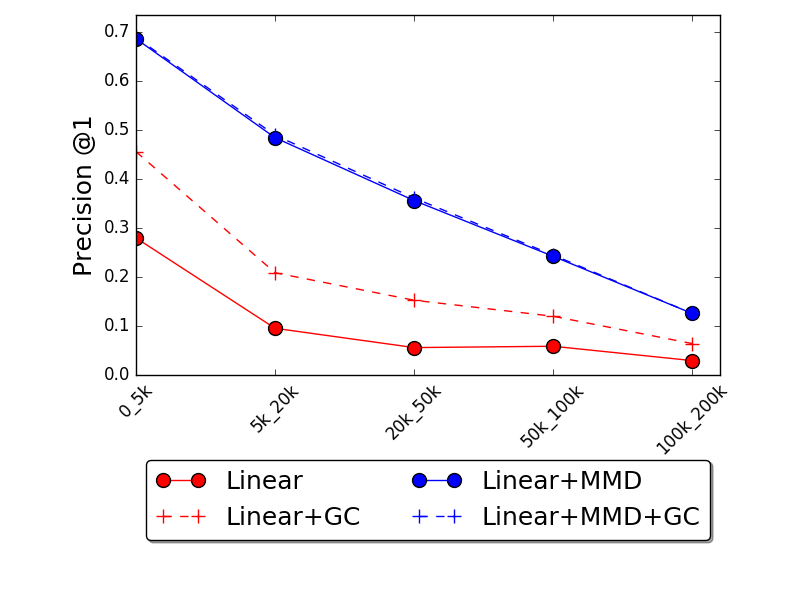}
    \end{subfigure}%
    \begin{subfigure}
        \centering
        \includegraphics[height=2.5in]{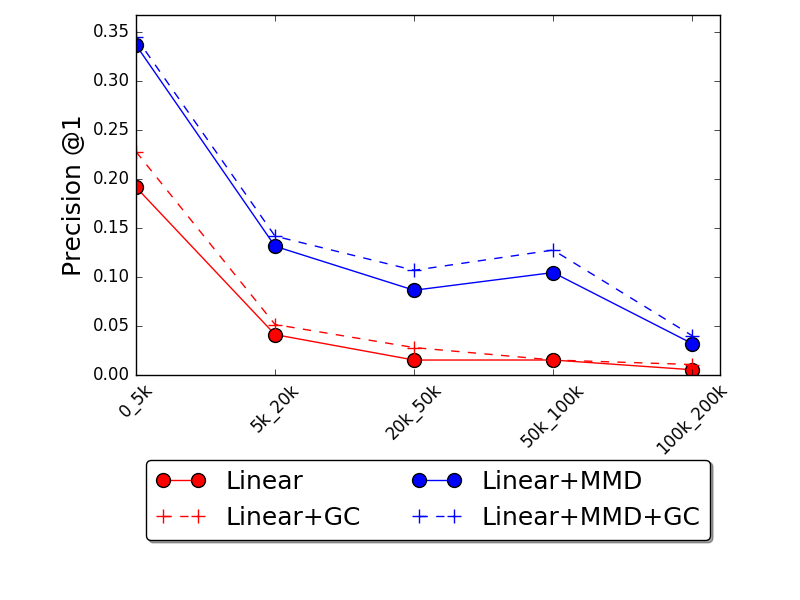}
    \end{subfigure}
    \caption{Model performance as a function of English word frequency bins using the top 5000 (left) and 750 (right) EN-IT word pairs as training data. Precision@1 refers to the fraction of words correctly translated by the method on held out testing sets.}
        \label{emb_comp}
\end{figure*}

Figure \ref{emb_comp} shows that the semi-supervised MMD-Net was able to significantly outperform the standard linear regression on a paired dataset of $750$ and $5000$ word-translation pairs in every frequency bin . Furthermore, this dominance over linear regression follows a similar pattern in the precisions @5 and @10. The method also outperformed several other linear and nonlinear methods as shown in Table \ref{table:comp}.

\begin{table*}[t]
	\caption{Comparison of Precision@1 across different algorithms and dimensionality reduction schemes. PCA S and PCA T refers to projecting the source and target respectively onto their first 270 principal vectors. KR refers to Kernel Ridge Regression an RBF refers to the radial basis function kernel with heuristically set scale}
    \vskip 0.15in
    \begin{center}
	\begin{tabular}{l|r|r|r|r|r}
        \hline
        \abovespace\belowspace
		{} &      0-5k &    5k-20k &   20k-50k &  50k-100k &  100k-200k \\
		\hline
        \abovespace
		Linear                 &  0.228 &   0.052 &    0.028 &     0.015 &      0.011 \\
		Linear + PCA S         &  0.236 &   0.057 &    0.031 &     0.036 &      0.019 \\
		Linear + PCA T         &  0.207 &   0.044 &    0.031 &     0.028 &      0.011 \\
		Linear + PCA S + T &  0.212 &   0.072 &    0.033 &     0.043 &      0.029 \\
		Random Forrest         &  0.008 &   0.000 &    0.000 &     0.000 &      0.000 \\
		KR 2-deg Poly          &  0.057 &   0.003 &    0.008 &     0.010 &      0.008 \\
		KR 3-deg Poly          &  0.049 &   0.005 &    0.003 &     0.013 &      0.008 \\
		KR RBF                 &  0.057 &   0.003 &    0.010 &     0.010 &      0.008 \\
				\belowspace
		Linear + MMD           &  \textbf{0.347} &   \textbf{0.129} &    \textbf{0.099} &     \textbf{0.094} &      \textbf{0.035} \\
        
		\hline
	\end{tabular}
    \end{center}
	\label{table:comp}
\end{table*}

\section{Discussion and Future Work}

The addition of the MMD cost function term significantly improves the results of regression in the low data regime. Furthermore, to the best knowledge of the authors, this method achieves state of the art results on the embeddings of \cite{lt_trans}. The authors also experimented with deeper nets, but did not observe significant performance improvements, an observation consistent with the observations of \cite{google_linear}.

\subsection{Adversarial Distribution Matching}
One promising future direction involves replacing the MMD unsupervised term with a Generative Adversarial Network (GAN) \cite{GAN}. Like the MMD, the GAN also involves a maximization over a function class of a measure of dissimilarity. Similarly, the GAN loss function can be used for unsupervised learning of probability distributions. However, the GAN is usually optimized directly by stochastic gradient descent, trading the quadratic time dependence on minibatch size with a linear one. In practice however, the maximization over the function class (the discriminator) is usually done in $k$ gradient descent steps for every one step of training the distribution matching net (the generator). Furthermore, the GAN cost function does not have a dependence on kernel scale. 

Analogous to the discriminator in the GAN, we can also adversarially learn the MMD. In this setup, the function class takes the the form of a parametrized network. Instead of estimating the supremum of the mean discrepancy over a ball in RKHS, we would be finding the supremum through gradient ascent on the network. This would also have the effect of eliminating the quadratic compute and the dependence on kernel scale. This formulation of the MMD would allow for a more direct comparison between the GAN and MMD loss functions, and warrants future investigation. These two loss functions are in-equivalent, as the only intersection between $f$-divergences, like the Jensen-Shannon Divergence which is equivalent to the GAN, and integral measures like the MMD is the total variation distance. \cite{mohamed2016learning} Thus, one might be able to leverage more diverse information by combining the two. 

\subsection{Bi-Directional Networks}
In the case of translation between two spaces of equal dimension, the inverse of the translation transformation should also be a translation from the target to the source space. We can capitalize on this observation to further constrain our set of possible translations. This allows the transformation to also draw information from the structure of the source space. More specifically one can minimize:

\begin{multline}
L =  \alpha_{target}\| RT - S \|_{target}^2 + \\ (1-\alpha_{target})\|R-ST^{-1}\|_{source}^2
\end{multline}

where $T \in GL_d$, $\alpha_{target} \in [0,1]$ and $R,S \in \R^{d\times n_{pair}}$. This would result in twice as much supervisory signal and maintain the same number of parameters. Furthermore, this can also be applied in conjunction with the GAN loss. It is also compatible with the pre-initialization scheme. In the case of a more complex nonlinear network where an inverse transformation cannot be easily calculated, the architecture could include an encoder network which maps from the source to the target and a decoding network which maps from the target to the source. These two mappings could then be constrained to be close to mutual inverses through a reconstruction loss penalty.

\bibliography{example_paper}
\bibliographystyle{icml2017}

\end{document}